\documentclass[10pt,twocolumn,letterpaper,table]{article}
 
\usepackage[algorithms]{wacv}
\def\wacvPaperID{1246} 
\def\confName{WACV}
\def\confYear{2025}

\usepackage[table]{xcolor}
\usepackage{graphicx}
\usepackage{booktabs}
\usepackage[pagebackref,breaklinks,colorlinks]{hyperref}
\usepackage{enumitem}
\usepackage{orcidlink}
\usepackage{url}
\usepackage{graphicx}
\usepackage{amsmath}
\usepackage{amssymb}
\usepackage{bbm}
\usepackage{float}
\usepackage{booktabs, makecell, tabularx}
\usepackage{algorithm}
\usepackage{algorithmicx}
\usepackage{algpseudocode}
\algrenewcommand\alglinenumber[1]{\tiny #1:}
\usepackage{footnote}
\usepackage{tablefootnote}
\usepackage{adjustbox}
\usepackage{newfloat}
\usepackage[utf8]{inputenc} 
\usepackage[T1]{fontenc}
\usepackage{listings}
\usepackage{bibentry}
\usepackage{arydshln}
\usepackage{pifont}
\usepackage{babel}
\usepackage{multirow}
\usepackage[font=small,labelfont=bf]{caption}
\usepackage[pangram]{blindtext}
\usepackage{wrapfig,lipsum,booktabs}
\usepackage{amsmath}
\usepackage{amsthm}
\usepackage{amsfonts,amsmath,amssymb}       
\usepackage{nicefrac}       
\usepackage{lipsum}
\usepackage{wasysym}
\usepackage{tikz}
\usepackage{subcaption}
\usepackage{cuted}

\newcommand\method{IEE}

\newcommand{\cmark}{\ding{51}}%
\newcommand{\xmark}{\ding{55}}%
\definecolor{lgray}{RGB}{245,245,245}
\definecolor{lgreen}{RGB}{236, 255, 201}
\definecolor{nvgreen}{RGB}{118, 185, 0}
\definecolor{cvprblue}{rgb}{0.21,0.49,0.74}

\usepackage[capitalize]{cleveref}
\crefname{section}{Sec.}{Secs.}
\Crefname{section}{Section}{Sections}
\Crefname{table}{Table}{Tables}
\crefname{table}{Tab.}{Tabs.}

\begin{document}
\sloppy
\title{Advancing Weight and Channel Sparsification with Enhanced Saliency} 


\author{Xinglong Sun, Maying Shen, Hongxu Yin, Lei Mao, Pavlo Molchanov, Jose M. Alvarez\\
{\small NVIDIA}\\
}

\maketitle
\begin{strip}
  \vspace{-1.5cm}
\begin{center}
\includegraphics[width=\linewidth]{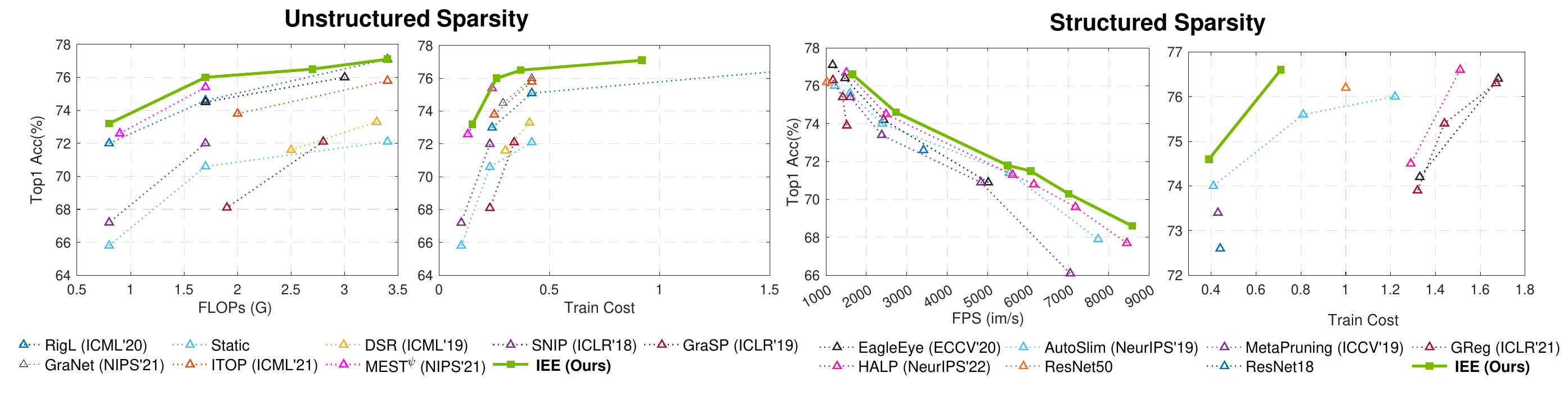}
\end{center}
\vspace{-0.35cm}
  \captionof{figure}{Our unstructured and structured pruning results on ImageNet1K. \textbf{Left:} Unstructured weight sparsity with different pruning ratios as a function of FLOPs and train cost, the top-left is better; \textbf{Right:} Structured pruning targeting various latency constraints, as a function of frame per second during inference where the top-right is better, and train cost the top-left is better. NVIDIA Titan V GPU is used to measure FPS. Train costs are reported as scales w.r.t training a baseline ResNet50.
  }
\label{fig:teaser}
\end{strip}
\begin{abstract}
Pruning aims to accelerate and compress models by removing redundant parameters, identified by specifically designed importance scores which are usually imperfect. This removal is irreversible, often leading to subpar performance in pruned models. Dynamic sparse training, while attempting to adjust sparse structures during training for continual reassessment and refinement, has several limitations including criterion inconsistency between pruning and growth, unsuitability for structured sparsity, and short-sighted growth strategies. Our paper introduces an efficient, innovative paradigm to enhance a given importance criterion for either unstructured or structured sparsity. Our method separates the model into an active structure for exploitation and an exploration space for potential updates. During exploitation, we optimize the active structure, whereas in exploration, we reevaluate and reintegrate parameters from the exploration space through a pruning and growing step consistently guided by the same given importance criterion. To prepare for exploration, we briefly "reactivate" all parameters in the exploration space and train them for a few iterations while keeping the active part frozen, offering a preview of the potential performance gains from reintegrating these parameters. We show on various datasets and configurations that existing importance criterion even simple as magnitude can be enhanced with ours to achieve state-of-the-art performance and training cost reductions. Notably, on ImageNet with ResNet50, ours achieves an $+1.3$ increase in Top-1 accuracy over prior art at $90\%$ ERK~\cite{mocanu2018scalable} sparsity. Compared with the SOTA latency pruning method HALP~\cite{shen2021halp}, we reduced its training cost by over $70\%$ while attaining a faster and more accurate pruned model.   
\end{abstract}

\section{Introduction}\label{sec:intro}
Contemporary advancements in deep learning for computer vision tasks~\cite{he2016deep, liu2016ssd, dosovitskiy2020image, xie2021segformer, sun2024refining, sun2023revisiting} have largely hinged on the development of deep neural networks (DNNs). As the literature progresses for improved performance, so do the model size, computation, and inference latency, which hinder deployment to applications that suffer stringent resources, \textit{e.g.,} edge device applications. To address this, \textbf{\textit{pruning}} has emerged as a crucial area of research. Pruning aims to induce sparsity in DNNs, either through structured (channel removal)~\cite{li2017pruning, molchanov2019importance, lin2020hrank, shen2021halp, sun2024multi} or unstructured approaches (weight removal)~\cite{alvarez2016learning, han2015deep, gale2019state,louizos2018learning, sun2022disparse}, for more efficient inference.

A key aspect of pruning research has been developing an \emph{importance score} to identify and remove the least important parameters. Various approaches have been introduced for calculating this importance score either for channels~\cite{li2017pruning, chin2020towards, he2020learning, he2018soft, yang2018netadapt, molchanov2019importance, lin2020hrank, shen2021halp, he2019filter, sun2024multi, sun2023pruning} or weights\cite{alvarez2016learning, han2015deep, gale2019state,louizos2018learning, sun2022disparse, zhou2021effective, srinivas2017training, molchanov2017variational}. Though yielding promising results, these scores are not as perfect as the Oracle choices~\cite{ding2019approximated, molchanov2016pruning}, \textit{which can only be obtained through exhaustive and time-consuming ablations}. Once the parameters are removed based on these imperfect scores, we can not recover them to reassess and potentially identify a better sparse structure. This results in unsatisfactory performance of the pruned models.

Towards addressing the above issue, a recent development called \textbf{\textit{dynamic sparse training}}~\cite{mocanu2018scalable, dai2019nest, evci2020rigging, liusw2021sparse, yuan2021mest, liu2021we, yuan2020growing, zhou2021efficient, hou2022chex} like the popular RigL~\cite{evci2020rigging}, aims to adjust sparse structures during training using a prune-and-growth approach, continually reassessing and refining the structure. Unlike pruning, the growth phase aims to reincorporate parameters that may have been prematurely pruned. However, these methods face several limitations: \textbf{(i)} they employ a unique growth importance criterion that differs from the pruning criterion, leading to a potential cycle where most newly grown parameters might be pruned again due to this criterion inconsistency, thus diminishing the effectiveness of the growth process; \textbf{(ii)} they are typically infeasible for structured sparsity due to their reliance on gradients over zeroed parameters which are simply zero following the chain-rule for zeroed channels; \textbf{(iii)} their growth strategies are usually short-sighted, focusing only on immediate impacts over the subsequent data batches, potentially limiting long-term effectiveness.

In this paper, we present an innovative procedure called \emph{\textbf{IEE}} to \emph{enhance the effectiveness of a given importance criterion for either unstructured or structured sparsity.} Our approach first divides the model into two sections: an active sparse structure and an exploration space containing inactive or pruned parameters. We then enhance the active structure through a process of \textbf{\textit{I}}terative \textbf{\textit{E}}xploitation and \textbf{\textit{E}}xploration. In the exploitation phase, we treat the current active structure as optimal, focusing on its training for convergence. During the exploration phase, we question this assumed optimality. Utilizing the provided importance score, we start by pruning less crucial parameters from the current sparse model. Subsequently, we briefly "reactivate" all parameters in the exploration space, training them for a few iterations while keeping the active part frozen. This step offers a preview of the potential performance gains from reintegrating these parameters. Finally, using the same importance score for consistency, we reassess and select the most vital parameters based on their readjusted weights from the exploration space, growing them into the active structure.
\begin{figure*}[t!]
\begin{center}
\includegraphics[width=0.8\linewidth]
{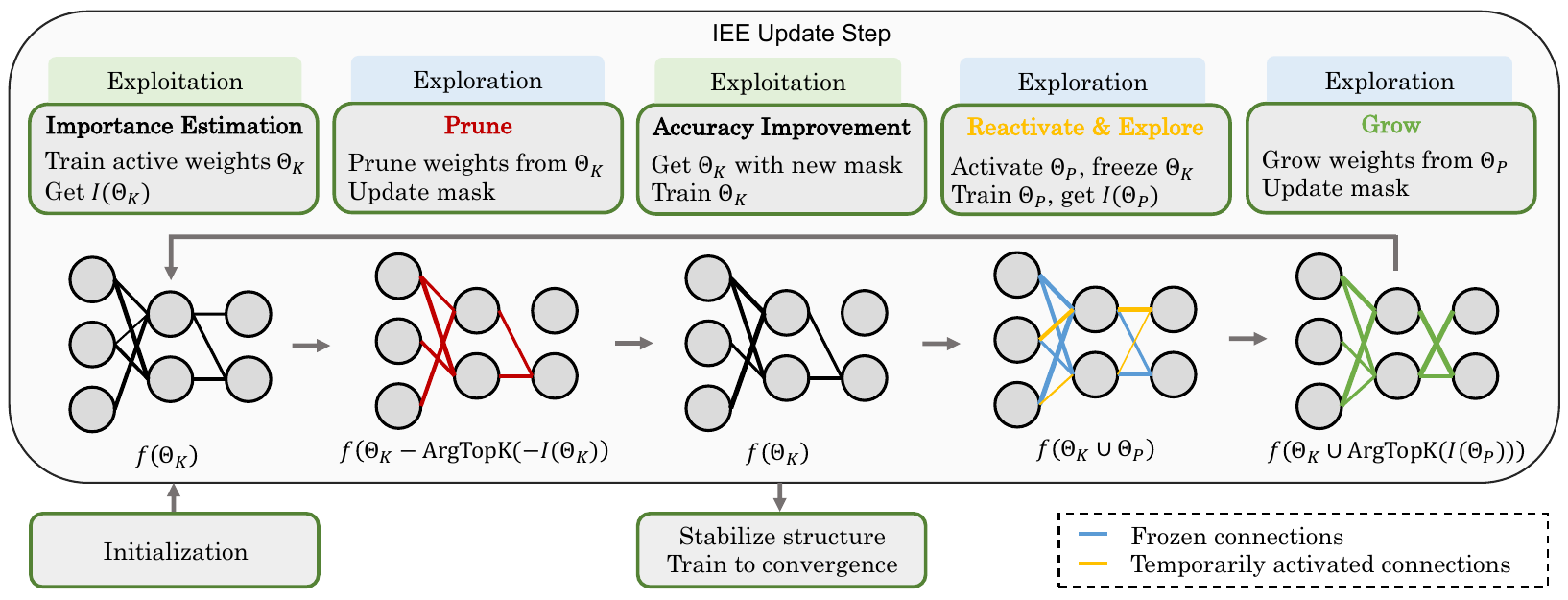}
\end{center}
\vspace{-0.35cm}
  \caption{Overview of our method. In each \method{} update step, we first train the active weights $\Theta_K$ for $H$ steps then prune a number of connections from $\Theta_K$. We later train the weights $\Theta_K$ just selected for $J$ steps for better exploiting the current architecture. To explore a potentially better sparse architecture, we temporarily activate the exploration space $\Theta_P$ and train them for $Q$ steps while freezing $\Theta_K$. We then evaluate the importance scores of the activated $\Theta_P$ to grow the top-ranked weights. This completes one full \method{} update step, and it is repeated until the update period ends.
  }
\label{fig:flowchart}
\vspace{-15pt}
\end{figure*}
We conduct comprehensive experiments to validate the effectiveness of our method on various datasets~\cite{krizhevsky2009learning,deng2009imagenet,everingham2010pascal} and models~\cite{he2016deep,zagoruyko2016wide,howard2017mobilenets,liu2016ssd} targeting both latency-aware structured sparsity and unstructured weight sparsity. We show that \textit{even saliency criterion simple as Magnitude~\cite{han2015deep} and Taylor~\cite{molchanov2019importance} scores} can be enhanced with ours to achieve state-of-the-art performance and surpass competitive pruning and sparse training methods by a margin. Notably, our method could either be applied to pretrained models or start from scratch, which yields \emph{substantial training cost savings}. Figure~\ref{fig:teaser} offers a glimpse of our comprehensive experiments. When targeting ResNet50 on ImageNet1K, compared with the best hardware-aware channel pruning approach HALP~\cite{shen2021halp}, our method consumes $\mathbf{70\%}$ less training cost ($\mathbf{\times0.39}$ v.s. $\times1.29$) while surpassing its performance with both higher FPS ($\mathbf{2736}$ v.s. $2621$) and Top-1 accuracy ($\mathbf{74.6}$ v.s. $74.5$). Compared with the popular RigL~\cite{evci2020rigging} for unstructured sparsity, our method yields a $\mathbf{+1.3}$ improvement in Top1 accuracy ($\mathbf{74.3}$ v.s. $73.0$) at $90\%$ ERK sparsity. Additionally, we introduced metrics to quantitatively assess effectiveness of sparse structure exploration, demonstrating our method's superior performance over dynamic sparse training techniques.

Our contributions can be summarized as follows:
\begin{itemize}
    \item We propose \method{}, a simple and effective paradigm to enhance importance score leveraging a new iterative exploitation-exploration feedback.
    \item We show on a variety of datasets and models for both structured and unstructured sparsity that importance criterion even simple as magnitude and Taylor scores~\cite{molchanov2019importance} can be enhanced with ours to achieve state-of-the-art results, accompanied by substantial reductions in training costs.
    \item We also design new metrics to quantitatively assess the effectiveness of sparsity structure exploration strategy and benchmark ours against previous dynamic sparse training approaches.
\end{itemize}

\section{Related Works}
\label{sec:related}
In general, our work is related to pruning and dynamic sparse training. We will now provide a brief recap of both of the two topics and highlight our differences.\\
\textbf{Pruning.}
A central focus in many pruning studies is developing a \emph{pruning importance score} to identify redundant parameters from the model for removal. The majority of pruning methods evaluate weight magnitude~\cite{thimm1995evaluating, strom1997sparse,han2015deep, narang2017exploring,zhu2017prune,gale2019state} on well-performed pretrained dense weights as importance scores. 
There have also been other importance criterion proposed such as Hessian-based scores~\cite{lecun1990optimal, hassibi1992second} and probability-based scores~\cite{zhou2021effective, srinivas2017training, molchanov2017variational}. In this paper, we study our method with weight magnitude metric~\cite{han2015deep} for unstructured sparsity. 

Some works, targeting structured sparsity, aim to prune convolutional filters~\cite{li2017pruning} or attention heads~\cite{michel2019sixteen}, thus enjoy immediate memory and computation benefit without specialized hardware and library support~\cite{han2016eie}. Exemplary channel importance criterion relied on metrics like weight norm~\cite{li2017pruning, chin2020towards, he2020learning, he2018soft, yang2018netadapt, sun2024towards}, Taylor expansion~\cite{lin2018accelerating, molchanov2019importance, you2019gate}, geometric median~\cite{he2019filter}, and feature maps rank~\cite{lin2020hrank}. Other works~\cite{chen2018constraint, shen2021halp, shen2023hardware} including the very recent HALP~\cite{shen2021halp} consider channel pruning under a latency or FLOPs constraint, aiming for more hardware-friendly structured sparse architecture with practical speed-up. In our structured sparsity experiment, we study our method with Taylor score~\cite{molchanov2019importance} and leverage the hardware-aware pruning scheme developed in HALP to directly achieve a sparse model optimized in inference latency.

\noindent\textbf{Dynamic Sparse Training.}
Albeit the decent performance of pruning on pretrained models, the dense model pretraining is usually computationally demanding and redundant. Moreover, once parameters are pruned, they cannot be recovered for reassessment and potential improvement of the sparse structure later in training. Towards addressing these goals,  a group of works~\cite{mocanu2018scalable, dettmers2019sparse, mostafa2019parameter, kusupati2020soft, wortsman2019discovering, dai2019nest, evci2020rigging, lin2020dynamic, ma2021effective, yuan2020growing, liusw2021sparse, yuan2021mest, liu2021we, yuan2022layer, hou2022chex, lasby2023dynamic}, usually referred to as \textit{Dynamic Sparse Training}, have considered the idea of repeated alternating prune-grow sessions to dynamically configure the sparsity structure through training from scratch, giving the model more flexibility. A key aspect of these approaches is the development of a \emph{growing importance score} to identify and grow back the prematurely pruned parameters. For instance, RigL, a pivotal study in this field, uses magnitude-based pruning but suggests regrowing parameters based on immediate gradients of zeroed and pruned weights, greedily optimizing their effectiveness for the subsequent data batch's gradient descent. Building on RigL's success, various adaptations have emerged, such as starting with lower sparsity~\cite{liusw2021sparse} or smaller batch size and longer update intervals~\cite{liu2021we}, or incorporating layer freezing and data sieving techniques\cite{yuan2021mest, yuan2022layer}, aiming to boost accuracy or reduce training costs. Despite advancements, the prevalent reliance on greedy exploration strategy undermines the long-term quality of the resulting model. Moreover, the discrepancy in the criteria for pruning and growing could lead to a suboptimal exploration of new architectures, where newly grown parameters are mostly pruned before being fully exploited.

Implementing structured sparsity in dynamic sparse training has been minimally addressed in prior work due to the challenge that gradients over pruned channels lead to zero values following chain-rule in backpropagation, making previous growth criteria infeasible. Existing solutions~\cite{he2018soft, kang2020operation, lym2019prunetrain, yuan2020growing, lasby2023dynamic} use alternatives to prune-grow, such as soft-masking~\cite{he2018soft, kang2020operation} and group-lasso regularization~\cite{lym2019prunetrain}. While the recent SCS~\cite{yuan2020growing} incorporates structured parameter exploration into optimization and uses continuous sparsification, it still relies on dense gradients throughout training, limiting its practicality and the possibility of a sparsified backward pass.

Like many dynamic sparse training approaches, our method also employs a prune-grow dynamics to update the structure. However, a key distinction is that we use the same given importance criterion for both pruning and growing parameters for either unstructured or structured sparsity without dense gradients reliance during the training process. 

\section{Methodology}
\label{sec:method}

\begin{algorithm}[t]
    \caption{\method{} Pseudocode}\label{euclid}
    \label{algo:1}
    \textbf{Input:}$\Theta$: Target Model, $\mathcal{D}$: Dataset, $\Psi$: Latency Constraint, $T$: Total \method{} Update Steps, $\Omega^t$: \method{} Update Budget, $H$, $J$, $Q$: Number of Iterations for our \textit{Importance Estimation}, \textit{Accuracy Improvement}, and \textit{Reactivate \& Explore} stages.
    \begin{algorithmic}[1]
        \State{Initialialize $\Theta_K, \Theta_P$ such that $Z(\Theta_K) = \Psi$}
        \State{$\Delta T \gets H+J+Q$}
        \State{$t, flag \gets 0, 0$}
        \For{$i \gets 1$ to $|\mathcal{D}|$}
            \If{$(i+J+Q) \mod \Delta T = 0$ and $t < T$} 
            \State{Collect importance score $I(\Theta_K)$}
            \State{Perform Prune according to Eqn.~\ref{eqn:lookahead_prune}}
            \ElsIf{$(i+Q) \mod \Delta T = 0$ and $t < T$}
            \State{$flag \gets 1$}
            \ElsIf{$i \mod \Delta T = 0$ and $t < T$}
            \State{Collect importance score $I(\Theta_P)$}
             \State{Perform Grow according to Eqn.~\ref{eqn:lookahead_grow}}
            \State{$t \gets t + 1$}
            \State{$flag \gets 0$}
            \EndIf
            \If{$flag$}
            \State{\texttt{//Reactivate \& Explore}}
            \State{Train $\Theta_P$ with $\Theta_K$ frozen according to Eqn.~\ref{eqn:reexp}}
            \Else{}
            \State{\texttt{//Importance Estimation}}
            \State{\texttt{//Accuracy Improvement}}
            \State{Train $\Theta_K$ according to Eqn.~\ref{eqn:impest}}
            \EndIf
        \EndFor
    \end{algorithmic}
\end{algorithm}
\noindent\textbf{Notation}. 
Let us consider a neural network with weights $\Theta$. We separate the entire weights into two parts: the \emph{currently active sparse structure $\Theta_K$}, and \emph{an exploration space $\Theta_P$ containing inactive or previously pruned weights}. $\Theta_K$ is also going to be our final selected structure. Moreover, suppose we are given a computational resource budget $\Psi$ for the pruned model and an importance score $I(.)$ to measure parameter saliency. Without loss of generality, for \emph{unstructured sparsity}, $\Psi$ is directly the target sparsity level, and we consider $I(.)$ to be the magnitude score~\cite{han2015deep}. For structured sparsity, we align $\Psi$ directly with the forward inference latency of the model and consider $I(.)$ to be the Taylor score~\cite{molchanov2019importance}. To enable inference latency reduction, we leverage the hardware-aware pruning approach HALP~\cite{shen2021when} and detail our integration with it in the Appendix. We further declare a function $\mathcal{R}(\cdot)$ such that $\mathcal{R}(\Theta_K)$ returns the current computation resource exploited by active structure $\Theta_K$. Finally suppose we are given a training dataset $\mathcal{D}$ consisting of $N$ input-output samples $\{(x_i,y_i)\}_{i=1}^N $ and a training loss function $\ell(\cdot)$.

\noindent\textbf{Method Description.} Our method begins by randomly initializing $\Theta_K$ and $\Theta_P$ such that $\mathcal{R}(\Theta_K) = \Psi$. During training, we enhance $\Theta_P$ by iteratively exploiting and exploring the sparse structure, guided by the given importance criterion $I(.)$. Each exploitation-exploration cycle involves training the active structure $\Theta_K$ for maximizing performance and updating it for structure improvement with a prune-grow. We perform a total of $T$ such cycles repetitively, concluding at $3/4$ of the total training iterations. To manage the proportion of weights updated in each cycle, we introduce $\Omega^t$ as an "update budget," limiting the number of parameters to be pruned and grown at each $t$-th \method{} step. Initially set at $\Omega^0 = 0.3\Psi$, it gradually decreases, guided by a scheduler~\cite{de2020progressive} as $t$ increases. Next, we will discuss the details of each exploitation-exploration step for structure update below.

\subsection{Our Update Step}
\label{subsec:lookahead_method}
Each \method{} step consists of five stages: \textit{Importance Estimation}, \textit{Prune}, \textit{Accuracy Improvement}, \textit{Reactivate \& Explore}, and \textit{Grow}. The overview of the method is shown in Figure~\ref{fig:flowchart} and the algorithmic description in Algorithm~\ref{algo:1}. 

\noindent\textbf{Importance Estimation.} The commencement of the \method{} step involves \emph{exploiting} the active sparse structure $\Theta_K$ to maximize its performance. During this phase, $\Theta_K$ undergoes $H$ iterations of training, wherein the importance of each parameter $I(\Theta_K)$ is also assessed. This training can be formulated as:
\begin{equation}
    \label{eqn:impest}
    \min_{\Theta_K} \sum_{i=1}^{H} \ell(f({\Theta_K}; \mathbf{x}^i), \mathbf{y}^i).
\end{equation}
\textbf{Prune.} Later, to initiate an update to the active sparse structure for \emph{exploring} a new potentially better one, we remove a portion of redundant parameters from $\Theta_K$ based on the importance score $I(\Theta_K)$. The pruned parameters at the $t$-th \method{} update step can be denoted as $\text{ArgTopK}(-I(\Theta_K), \Omega^t)$, which are the parameters with the least importance that still comply with the update budget $\Omega^t$. The pruned parameters are then added to exploration space $\Theta_P$ for future potential revival. These can be  formulated as:
\begin{align}
\label{eqn:lookahead_prune}
    &\Theta_K \gets \Theta_k - \text{ArgTopK}(-I(\Theta_K), \Omega^t)\\
    &\Theta_P \gets \Theta_P + \text{ArgTopK}(-I(\Theta_K), \Omega^t)\nonumber
\end{align}
After pruning, computation resource of active sparse structure $\Theta_K$ is reduced by $\Omega^t$ from $\Psi$, i.e. $\mathcal{R}(\Theta_K) = \Psi - \Omega^t$. Moreover, with unstructured sparsity, parameters are directly removed solely based on their importance to achieve the target sparsity. In the structured sparsity case, we also consider the latency cost and leverage a Knapsack solver following HALP~\cite{shen2021halp} to select the least important channels conforming to the latency constraint. 

\noindent\textbf{Accuracy Improvement.} With the new $\Theta_K$, we carry out another training stage for $J$ iterations to \emph{exploit} it with the goal of stabilizing it and improving its performance. This training stage can be formulated the same as Eqn.~\ref{eqn:impest}. We later show in ablation that this further exploitation step is crucial to the performance of \method{}.

\noindent\textbf{Reactivate \& Explore}.
We now \emph{explore} the exploration space $\Theta_P$ to identify parameters that may have been prematurely pruned yet still have the potential to enhance model performance. For a solid exploration, we temporarily re-activate all of the potential parameters in $\Theta_P$ then quickly update them for $Q$ iterations while $\Theta_K$ remains \emph{frozen}. We also leverage the given importance criterion to evaluate the saliency for these revived parameters $I(\Theta_P)$, which offers a preview of how these parameters might impact performance if reintegrated into $\Theta_K$. We later show in ablation that this freezing of $\Theta_K$ is crucial to the performance by preserving the currently selected architecture for a stable exploration. Moreover, when reactivated, $\Theta_P$ inherit their \textbf{MRU (\textit{i.e.,} Most Recently Used values)} before they were turned off as initialization for this $Q$ steps of training. This step can be formulated as:
\begin{equation}
    \label{eqn:reexp}
    \min_{\Theta_P} \sum_{i=1}^{Q} \ell(f({\Theta_P \cup \Theta_K}; \mathbf{x}^i), \mathbf{y}^i).
\end{equation}

Our exploration technique here also draws parallels to Optimistic Initialization strategies~\cite{machado2015domain, lobel2022optimistic} commonly utilized in Reinforcement Learning to \textit{mitigate greediness in exploration} by initially assuming all actions to be optimal then challenging this assumption by exploring them at least a few times. In our scenario, an 'action' refers to the reintroduction of a certain set of parameters from $\Theta_P$. Unlike dynamic sparse training methods like RigL, which greedily grow parameters based on immediate gradients, we offer all potential parameters an opportunity to demonstrate their effectiveness in conjunction with the current active structure $\Theta_K$ through this short period of revival.

\noindent\textbf{Grow.} The \emph{exploration} phase concludes with a growth process, finalizing the update to the active structure $\Theta_K$. The newly collected Importance score $I(\Theta_P)$ allows us to choose which parameters may enhance the performance. Therefore, at the $t$-th \method{} update step, we simply grow the parameters given by $\text{ArgTopK}(I(\Theta_P), \Omega^t)$, which are the highest importance parameters that satisfy the update budget $\Omega^t$. The grown parameters are then transferred from $\Theta_P$ to $\Theta_K$. These can be formulated as:
\begin{align}
\label{eqn:lookahead_grow}
    &\Theta_K \gets \Theta_k + \text{ArgTopK}(I(\Theta_P), \Omega^t) \\
    &\Theta_P \gets \Theta_P - \text{ArgTopK}(I(\Theta_P), \Omega^t)\nonumber
\end{align}

Similar to pruning, in the structured sparsity case, we leverage a knapsack solver as in HALP~\cite{shen2021halp} to determine the grown channels under latency constraint. It is noteworthy that the computational resource budget of the model remains constant before and after one \method{} step, maintaining the target value $\Psi$ ($\mathcal{R}(\Theta_K) = \Psi$).  We then cycle back to \emph{Importance Estimation} for another \method{} step until we complete all of the $T$ steps.

\section{Experiments}
\label{sec:exps}
We next demonstrate the effectiveness of our approach
across a comprehensive set of scenarios. We first study our method with \textit{structured sparsity} with \textit{Taylor importance criterion}~\cite{molchanov2019importance} for ResNet50 and MobileNet-V1~\cite{howard2017mobilenets} on ImageNet1K~\cite{deng2009imagenet}. To show the generability of \method{}, we also include object detection results on PASCAL VOC~\cite{everingham2010pascal}. We then study our method with \textit{unstructured sparsity} using \textit{magnitude importance criterion}~\cite{han2015deep} on ResNet50 and WideResNet22-2~\cite{zagoruyko2016wide} for ImageNet1K and CIFAR10~\cite{krizhevsky2009learning}. We observe that importance criterion simple as \textit{Taylor} scores~\cite{molchanov2019importance} and magnitude~\cite{han2015deep} can be enhanced with ours to achieve state-of-the-art results. Finally, we ablate our method and analyze the effectiveness of \method{} exploration strategy. We run the experiments on 8 Nvidia Tesla V100 GPUs for ImageNet1K and 1 GPU for CIFAR-10. We include more details in the Appendix.

\subsection{Structured Sparsity}
For our structured experiments, we leverage the hardware-aware compression framework proposed in HALP~\cite{shen2021halp} to directly reduce the inference latency of the model. We show that our method can enhance the importance criterion working both from scratch and on pretrained models. We compare with various competitive methods including prior art on resource or hardware-aware pruning like~\cite{li2020eagleeye, liu2019metapruning, shen2021halp}, standard channel pruning methods like ~\cite{tang2020scop, zhuang2020neuron}, dynamic sparse training method SCS~\cite{yuan2020growing}, and AutoML and NAS-based methods such as AutoSlim~\cite{yu2019autoslim} and AMC~\cite{he2018amc}. Our evaluation focuses on Top-1 and Top-5 accuracy and frames per second (FPS) to assess pruning effectiveness. Additionally, we provide \textbf{\textit{Train FLOPs}} to illustrate how our method effectively reduces training costs. Train FLOP calculations are provided in the Appendix.

\begin{figure}[!t]
\begin{center}
   \includegraphics[width=\linewidth]{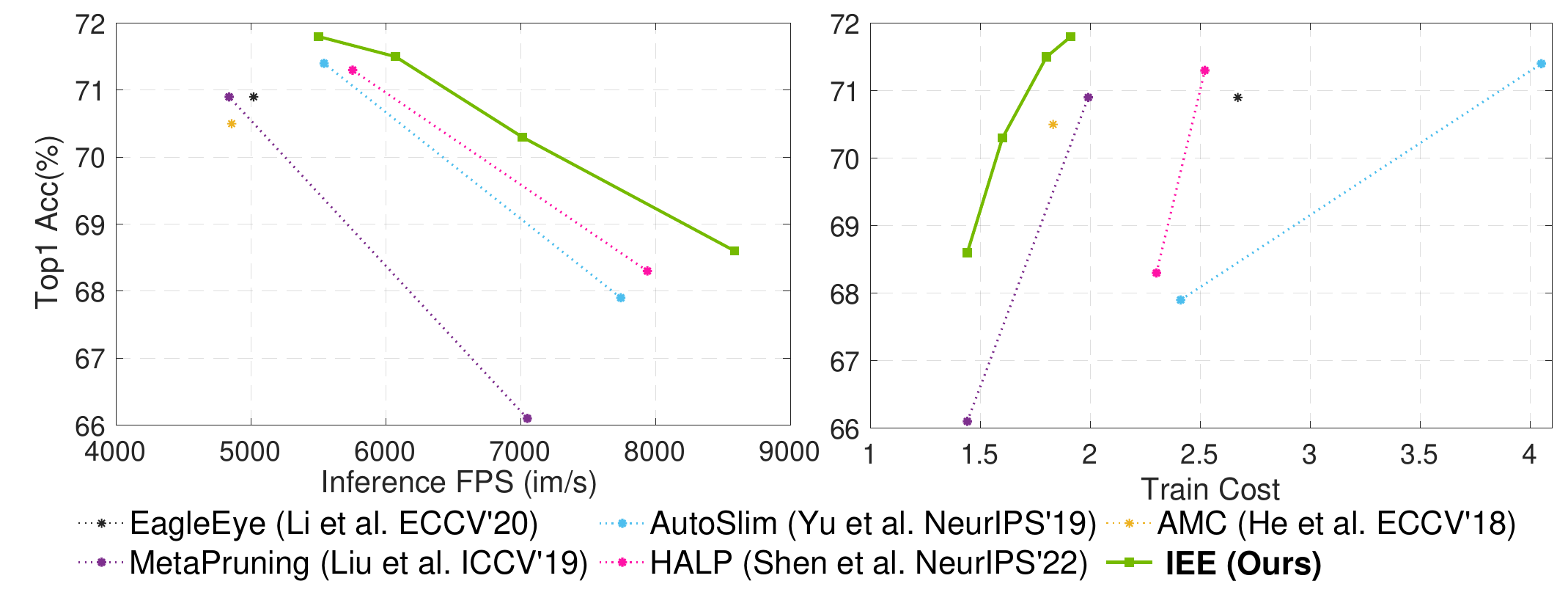}
\end{center}
    \vspace{-0.4cm}
   \caption{\textbf{ImageNet1K} structured sparsity results on MobileNet-V1 as a function of FPS (left, top-right is better) and training cost (right, top-left is better). FPS is measured on NVIDIA Titan V GPU; training costs are reported relative to dense MobileNet-V1.}
   \label{fig:mobilenet}
    \vspace{-0.6cm}
\end{figure}

\begin{table*}[t!]
    \centering
    \vspace{-0.2cm}
    \resizebox{\textwidth}{!}
    {
        \begin{tabular}{lcccccc}
            \toprule
            \rowcolor{lgray} 
            \textsc{Method} & \textsc{Top-1}($\%$)$\uparrow$ & \textsc{Top-5}($\%$)$\uparrow$  & \textsc{FLOPs}($\times e^9$)$\downarrow$ & \textsc{FPS(im/s)}$\uparrow$ & \textsc{EPOCHS}
            & \textsc{Train FLOPs($\times e^{18}$)$\downarrow$}
            \\
            \midrule
            \textsc{Dense}~\cite{li2020eagleeye} & $77.2$ &$92.9$ & $4.1$&$1019$&$90$&$\times1 \ (\text{w.r.t.} 1.6)$\\
            \midrule
            \multicolumn{7}{c}{\textsc{\textbf{Applied on Pretrained Model}}}\\
            \textsc{EagleEye}-2G\cite{li2020eagleeye} & $76.4$ &$92.9$ & $2.1$ & $1471$ &$90+120$&$\times 1.68$\\
            \textsc{GReg-2}\cite{wang2021neural} & $75.4$ & --& $1.8$ & $1414$ &$90+90$ &$\mathbf{\times1.44}$\\
            \textsc{SCOP}\cite{tang2020scop} & $76.0$ & & $2.2$ & -- &$90+140$ &$\times 1.83$\\
            \textsc{GBN}\cite{you2019gate} & $76.2$ &$92.8$ & $2.4$ & -- &$90+260$ &$\times 2.69$\\
            \textsc{HALP-$55\%$}\cite{shen2021halp} & $76.6$ &$\mathbf{93.2}$ & $2.1$ & $\mathbf{1672}$ &$90+90$ &$\times1.51$\\
            \rowcolor{lgreen} \textbf{\textsc{\method{}-$\mathbf{55\%}$}} & $\mathbf{77.0\pm0.08}$ &$\mathbf{93.2\pm0.15}$& $2.0$ & $1554$ &$90+130$& $\times1.71$\\
            \hdashline
            \textsc{EagleEye-1G}\cite{li2020eagleeye} & $74.2$ &$91.8$ & $1.0$ & $2429$ &$90+120$ &$\times1.33$\\
            \textsc{GReg-2}\cite{wang2021neural} & $73.9$ & --& $1.3$ & $1514$ &$90+90$ &$\times1.32$\\
            \textsc{DSNet}\cite{li2021dynamic} & $74.6$ &-- & $1.2$ & -- &$90+150$ &$\times1.49$\\
            \textsc{Polarize}\cite{zhuang2020neuron} & $74.2$ &-- & $1.2$ & -- &$90+158$ &$\times1.51$\\
            \textsc{HALP-$30\%$}\cite{shen2021halp} & $74.5$ &$91.8$ & $1.2$ & $2597$ &$90+90$ &$\mathbf{\times1.29}$\\
            \rowcolor{lgreen} \textbf{\textsc{\method{}-$\mathbf{30\%}$}} & $\mathbf{74.8\pm0.04}$ &$\mathbf{92.2\pm0.26}$ & $1.1$ & $\mathbf{2621}$ &$90+130$ &$\times1.39$\\
            \midrule
            \midrule
            \multicolumn{7}{c}{\textsc{\textbf{Start from Scratch}}}\\
            DSA\cite{ning2020dsa} & $74.7$ &$92.1$ & $2.0$ & -- & $120$&$\times1.11$\\
            SCS\cite{yuan2020growing} & $75.2$ &-- & $2.1$ & -- & $120$&$\times1.12$\\
            \textsc{TAS}\cite{dong2019network} & $76.2$ &$\mathbf{93.1}$ & $2.3$ & -- &$240$ &$\times1.50^*$\\
            \textsc{SRigL}\cite{lasby2023dynamic} & $76.2$ &-- & $2.0$ & -- &$515$ &$\times2.87$\\
            \rowcolor{lgreen}
        \textbf{\textsc{\method{}-$\mathbf{55\%}$}} & $\mathbf{76.6\pm0.17}$ &$\mathbf{93.1\pm0.20}$ & $2.1$ & $\mathbf{1654}$ &$130$ & $\mathbf{\times0.71}$\\
            \hdashline
            
            \textsc{MetaPruning}\cite{liu2019metapruning} & $73.4$ &-- & $1.0$ & $2381$ &$160$ &$\times0.43^*$\\
            \textsc{DMCP}\cite{guo2020dmcp} & $74.1$ &-- & $1.1$ & -- &$150$ &$\times0.45^*$\\
            \textsc{SRigL}\cite{lasby2023dynamic} & $73.6$ &-- & $1.0$ & -- &$515$ &$\times1.97$\\
            \rowcolor{lgreen} \textbf{\textsc{\method{}-$\mathbf{30\%}$}} & $\mathbf{74.6\pm0.22}$ &$\mathbf{92.1\pm0.18}$& $1.0$ & $\mathbf{2736}$ &$130$ &$\mathbf{\times0.39}$\\
            \bottomrule
        \end{tabular}
    }
    \caption{\textbf{ImageNet1K} structured sparsity results using ResNet-50, averaged \textit{over two runs}. \method{}-X\% refers to the percentage of parameters remaining after training. Training flops calculation is in Appendix. Results are further grouped by FLOPs. "$*$": estimated lower-bound train cost for NAS-based methods. SRigL~\cite{lasby2023dynamic} focuses on N:M Ampere sparsity, and \textbf{\textit{\method{} with N:M sparsity can be found in Appendix}}.}
    \vspace{-0.5cm}
    \label{table:imageresnet50structured}
\end{table*}
\noindent\textbf{Results on ImageNet1K.} We show results for two different architectures namely ResNet50 and MobileNet-V1 on ImageNet1K dataset with update period of our approach as (H=J=Q=150) in Table~\ref{table:imageresnet50structured} and Figure~\ref{fig:mobilenet}. Our approach consistently outperforms all the referenced methods. For ResNet50, compared to the latest state-of-the-art latency pruning method HALP~\cite{shen2021halp} which also relies on Taylor importance~\cite{molchanov2019importance}, we surpass its performance($\mathbf{74.6}$ v.s. $74.5$ Top-1, $\mathbf{2736}$ v.s. $2597$ FPS) even by starting from scratch, which generates impressive train cost savings ($\mathbf{\times 0.39}$ v.s. $\times 1.39$). Compared to dynamic sparse training methods such as SCS~\cite{yuan2020growing} our approach yields up to $1.4$ ($\mathbf{76.7}$ v.s. $75.2$) accuracy improvement with a $2.1G$-FLOPs model. We can observe similar patterns for the more efficient MobileNet-V1 architecture in Figure~\ref{fig:mobilenet}, with \method{} obtaining a much more superior latency-accuracy tradeoff while consuming much less training cost.

\begin{figure}[!ht]
\begin{center}
   \includegraphics[width=\linewidth]{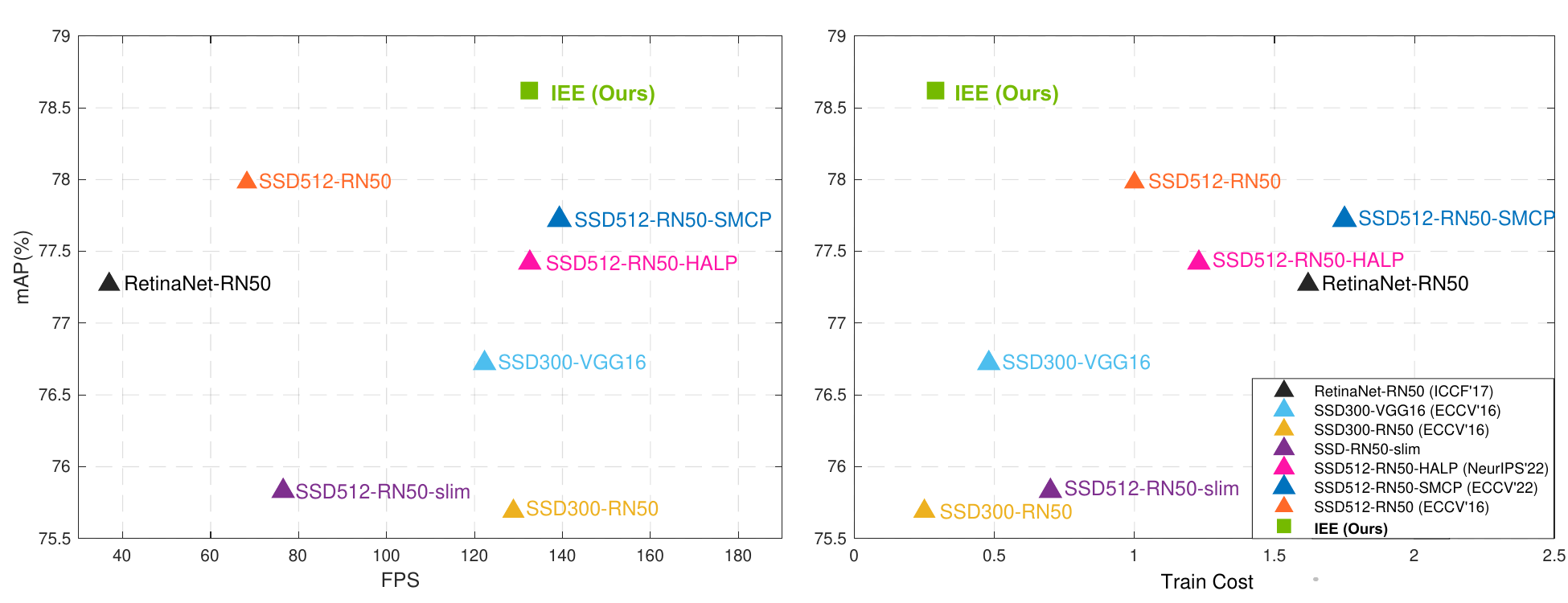}
\end{center}
\vspace{-0.5cm}
   \caption{\textbf{PASCAL VOC} structured sparsity results on SSD512-RN50 as a function of FPS (left, top-right is better) and training cost (right, top-left is better). FPS is measured on NVIDIA Titan V GPU; training costs are reported relative to dense SSD512-RN50.}
   \label{fig:detection_cmp_with_cost}
   \vspace{-0.5cm}
\end{figure}

\noindent\textbf{Generalization to Object Detection.}
We further demonstrate the performance of our approach on object detection. We report results with SSD512~\cite{liu2016ssd} using a ResNet50 backbone on the popular PASCAL VOC dataset~\cite{everingham2010pascal} in Figure \ref{fig:detection_cmp_with_cost}. To further clarify, we apply \method{} with training the entire detector \textit{from scratch}, rather than using a pruned ResNet50. As shown, \method{} clearly outperforms other competitive methods with higher mAP at faster or similar speed(FPS) and even surpasses the dense model with doubled inference speed. Tremendous training cost can also be spotted similar as the ImageNet1K~\cite{deng2009imagenet} case.

\begin{table*}[t!]
    \centering
    \vspace{-0.25cm}
    \resizebox{\textwidth}{!}
    {
        \begin{tabular}{lc|ccc|ccc}
            \toprule
            \rowcolor{lgray} 
             {\textsc{Method}} & \textsc{Total} & \multicolumn{3}{c|}{\textsc{Sparsity Ratio} $80\%$} & \multicolumn{3}{c}{\textsc{Sparsity Ratio} $90\%$} \\
            \rowcolor{lgray} 
             & \textsc{Epochs} & \textsc{Top-1}($\%$)$\uparrow$ & \textsc{FLOPs}($\times e^9$)$\downarrow$ & \textsc{Train FLOPs}($\times e^{18}$)$\downarrow$ & \textsc{Top-1}($\%$)$\uparrow$ & \textsc{FLOPs}($\times e^9$)$\downarrow$ & \textsc{Train FLOPs}($\times e^{18}$)$\downarrow$\\
            \midrule
            \textsc{Dense}~\cite{li2020eagleeye} &  & $76.8$ & $8.2$ &$\times1(\text{w.r.t.} 3.2)$& $76.8$ & $8.2$&$\times1(\text{w.r.t.} 3.2)$\\
            \midrule
            \multicolumn{8}{c}{}\\
            \multicolumn{8}{c}{\textbf{\textsc{Uniform Unstructured Sparsity Distribution}}}\\
            \textsc{Static} & $100$ & $70.6$ & $1.7$ &$\times0.23$& $65.8$ & $0.8$&$\times0.10$ \\
            \textsc{SNIP}~\cite{lee2018snip} & $100$ & $72.0$ & $1.7$ &$\times0.23$& $67.2$ & $0.8$&$\times0.10$ \\
            \textsc{SET}~\cite{mocanu2018scalable} & $100$ & $72.9$ & $1.7$ &$\times0.23$& $69.6$ & $0.8$& $\times0.10$\\
            \textsc{RigL}~\cite{evci2020rigging} & $100$ & $74.6$ & $1.7$ &$\times0.23$& $72.0$ & $0.8$& $\times0.10$\\
            \rowcolor{lgreen} 
            \textbf{\textsc{\method{}}} & $100$ & $\mathbf{75.6\pm0.19}$ & $1.7$ &$\times0.26$& $\mathbf{73.0\pm0.27}$ & $0.8$ &$\times0.16$\\
            \midrule
            \midrule
            \multicolumn{8}{c}{}\\
            \multicolumn{8}{c}{\textbf{\textsc{ERK~\cite{mocanu2018scalable, evci2020rigging} Unstructured Sparsity Distribution}}}\\
            \textsc{Static} & $100$ & $72.1$ & $3.4$ &$\times0.42$& $67.7$ & $2.0$&$\times0.24$ \\
            \textsc{RigL}~\cite{evci2020rigging} & $100$ & $75.1$ & $3.4$ &$\times0.42$& $73.0$ & $2.0$ &$\times0.24$\\
            \rowcolor{lgreen} 
            \textbf{\textsc{\method{}$_{0.8\times}$}} & $80$ & $\mathbf{75.6\pm0.23}$ & $3.4$ &$\times0.37$ &$\mathbf{73.6\pm0.21}$ & $2.0$& $\times0.24$\\ 
            \textsc{SNFS}~\cite{dettmers2019sparse} & $100$ & $75.2$ & $3.4$ &$\times0.61$& $72.9$ & $2.0$ &$\times0.50$\\
            \rowcolor{lgreen} 
            \textbf{\textsc{\method{}}} & $100$ & $\mathbf{76.2\pm0.08}$ & $3.4$ &$\times0.45$ &$\mathbf{74.3\pm0.31}$ & $2.0$& $\times0.30$\\
            \hdashline
            \textsc{DCIL}~\cite{kim2021dynamic} & $100$ & $76.2$ & $-$ &$\times 1.80^*$& $75.3$ & $-$ &$\times1.75^*$\\
            \textsc{IP-FT}~\cite{wimmer2022interspace} & $200$ & $77.2$ & $-$ &$\times 1.55^*$& $75.8$ & $-$ &$\times1.38^*$\\
            \textsc{Top-KAST}~\cite{jayakumar2020top} & $100$ & $76.4$ & $3.4$ &$\times 0.98$& $75.5$ & $2.0$ &$\times 0.64$\\
            \rowcolor{lgreen} 
            \textbf{\textsc{\method{}}$_{2\times}$} & $200$ & $77.1$ & $3.4$ &$\times$ $0.92$&$75.7$ & $2.0$ &$\times0.60$\\
            \textsc{RigL}~\cite{evci2020rigging} & $500$ & $77.1$ & $3.4$ &$\times2.10$ &$76.4$ & $2.0$ &$\times1.23$\\
            \rowcolor{lgreen} 
            \textbf{\textsc{\method{}}$_{5\times}$} & $500$ & $\mathbf{77.8\pm0.08}$ & $3.4$ &$\times2.30$ &$\mathbf{76.8\pm0.03}$ & $2.0$ &$\times1.50$\\
            \midrule
            \midrule
            \multicolumn{8}{c}{\textbf{\textsc{Non-Uniform Unstructured Sparsity Distribution}}}\\
            \textsc{MEST}~\cite{yuan2021mest} & $100$ & $75.4$ & $1.7$ &$\times0.24$& $72.6$ & $0.9$ &$\times0.13$\\
            \rowcolor{lgreen}
            \textbf{\textsc{\method{}$_{uniInit}$}} & $100$ & $\mathbf{76.0\pm0.16}$ & $1.7$ &$\times0.26$& $\mathbf{73.2\pm0.22}$ & $0.8$ & $\times0.15$\\
            \hdashline
            \textsc{DSR}~\cite{mostafa2019parameter} & $100$ & $73.3$ & $3.3$ &$\times0.41$& $71.6$ & $2.5$ &$\times0.30$\\
            \textsc{ITOP}~\cite{liu2021we} & $100$ & $75.8$ & $3.4$ &$\times0.42$& $73.8$ & $2.0$ &$\times0.25$\\
            
            \textsc{GraNet(s = 0.5)}~\cite{liusw2021sparse} & $100$ & $76.0$ & $3.0$ &$\times0.42$& $74.5$ & $1.7$ &$\times0.29$\\
            \rowcolor{lgreen} 
            \textbf{\textsc{\method{}}$_{erk-init}$} & $100$ & $\mathbf{76.5\pm0.17}$ & $2.7$ &$\times0.37$ &$\mathbf{74.6\pm0.13}$ & $1.7$& $\times 0.26$\\
            \hline
        \end{tabular}
    }
    \caption{\textbf{ImageNet1K} unstructured sparsity results on ResNet50, averaged \textit{over two runs}. \method{}${X\times}$ scales the baseline training epochs (100) by $X$. \method{}${uniInit}$ and \method{}$_{erkInit}$ (Non-Uniform) refer to using uniform or ERK for initializing the sparsity distribution. Results are grouped by \textbf{\textit{Train FLOPs}}. MEST~\cite{yuan2021mest} employs dataset sieving and layer freezing for cost reduction. $*$: approximated cost with ERK.} 
    \label{table:imageresnet50}
    \vspace{-0.35cm}
\end{table*}

\subsection{Unstructured Sparsity}
We now show results of our method on unstructured sparsity on ImageNet1K and CIFAR-10 datasets at different sparsity levels, FLOPs, and \textbf{\textit{training costs}}. We compare with the widely-referenced work RigL~\cite{evci2020rigging}, its follow-up works MEST~\cite{yuan2021mest} and ITOP~\cite{liu2021we}, and prominent SOTA soft pruning works DCIL~\cite{kim2021dynamic} and Top-Kast~\cite{jayakumar2020top}, etc. For a fair comparison, we demonstrate results following a fixed and predefined layer sparsity distribution including \textit{Uniform} and \textit{ERK}, avoiding redistributing the sparsity across layers throughout training as \textit{Non-Uniform} results. In \textit{Uniform}, the sparsity of each layer is equal to the total sparsity throughout training. In \textit{ERK}, \textit{Erd\H{o}s-R\'enyi-Kernel} (ERK) formulation~\cite{mocanu2018scalable, evci2020rigging} is adopted to set sparsity for each layer. With \textit{Non-Uniform} sparsity, the model could be initialized with Uniform or ERK distribution, or completely random.

\noindent\textbf{Results on ImageNet1K.} For this experiment, we used ResNet50~\cite{he2016deep} and set the update period of our approach as (H=J=Q=150). Table~\ref{table:imageresnet50} compares the performance of \method{} with prior works under different configurations. As shown, our approach consistently outperforms all the other methods by a significant margin. For instance, compared to RigL~\cite{evci2020rigging} under 100 training epochs at ERK sparsity distribution, our approach yields an improvement of $1.1\%$ and $1.3\%$ at $80\%$ and $90\%$ sparsity, respectively. Compared with the latest Non-Uniform dynamic sparse training methods~\cite{liusw2021sparse, liu2021we, yuan2021mest} with  augmentation techniques like data sieving, \method{} still outperforms in terms of tradeoffs between Top-1, FLOPs, and training cost.  Moreover, at longer epochs, our approach achieves better or comparable accuracy with \emph{much less training cost} compared with the latest works~\cite{wimmer2022interspace, kim2021dynamic} and even yields a $1\%$ top-1 accuracy improvement at $80\%$ ERK sparsity trained for $500$ epochs compared to the dense ResNet50 baseline.

\begin{table}[t!]
    \centering
    \resizebox{.47\textwidth}{!}
    {
        \begin{tabular}{lc|cc|cc}
            \toprule            
            \rowcolor{lgray} 
            \textsc{Method} & \textsc{Total} & \multicolumn{2}{c}{\textsc{Sparsity Ratio} $80\%$} & \multicolumn{2}{c}{\textsc{Sparsity Ratio} $90\%$}\\
            \rowcolor{lgray} 
             & \textsc{Epochs} & \textsc{Top-1}($\%$)$\uparrow$ & \textsc{FLOPs}($\times e^8$)$\downarrow$ & \textsc{Top-1}($\%$)$\uparrow$ & \textsc{FLOPs}($\times e^8$)$\downarrow$\\
            \midrule
            \textsc{Dense}~\cite{zagoruyko2016wide} & & $94.6$ & $3.2$ & $94.6$ & $3.2$ \\
            \midrule
            \textsc{Pruning}~\cite{gale2019state} & $250$ & $93.5$ & $0.5$ & $93.3$ & $1.1$ \\
            \textsc{Static} & $250$ & $92.9$ & $0.5$ & $91.6$ & $1.1$ \\
            \textsc{RigL}~\cite{evci2020rigging} & $250$ & $93.5$ & $0.5$ & $92.9$ & $1.1$ \\
            \rowcolor{lgreen} \textbf{\textsc{\method{}}}& $250$ & $\mathbf{93.8\pm0.07}$ & $0.5$ & $\mathbf{93.6\pm0.02}$ & $1.1$ \\ \midrule
            \textsc{Static} & $500$ & $93.2$ & $0.5$ & $91.8$ & $1.1$ \\
            \textsc{RigL}\cite{evci2020rigging} & $500$ & $93.7$ & $0.5$ & $93.3$ & $1.1$ \\
            \rowcolor{lgreen} \textbf{\textsc{\method{}}$_{2\times}$}& $500$ & $\mathbf{94.5\pm0.01}$ & $0.5$ & $\mathbf{93.8\pm0.01}$ & $1.1$ \\
            \bottomrule
        \end{tabular}
    }
    \caption{\textbf{CIFAR-10} unstructured sparsity results using WideResNet22-2. Averaged results \textit{over three runs}.}
    \label{table:widecifar}
    \vspace{-20pt}
\end{table}

\noindent\textbf{Results on CIFAR10.} We additionally evaluate our approach on WideResNet22-2 and CIFAR-10. As shown in Table~\ref{table:widecifar}, our approach outperforms the existing approaches with only half the training time. As also happened for ImageNet1K, our approach with $80\%$ sparsity and $500$ training epochs achieves almost the same performance as the dense model baseline. The results show the efficacy of \method{} extends to small dataset as well, surpassing dynamic sparse training strategies like RigL~\cite{evci2020rigging}.

\subsection{Ablation Studies}
\label{subsec:ablation}
In this section, we perform ablations and conduct additional analysis to validate our design choices and provide observations for (i) the update period, (ii) freezing parameters (see \emph{Reactivate \& Explore} in Section\ref{subsec:lookahead_method}), (iii) the effect of the \emph{Accuracy Improvement} stage, (iv) growing criterion of the neurons, and (v) initialization of the grown neurons. For these experiments, we study on ResNet50 with $90\%$ ERK sparsity trained for $100$ training epochs from scratch on ImageNet1K.

\noindent\textbf{Sensitivity to the update period.} 
We first study the effect of the update period $H,J,Q$, which controls the balance between exploration and exploitation. A longer update period leads to more exploitation of the current selected sparse structure but fewer explorations of new ones. As shown in Table~\ref{table:ablation}, we observe intuitive degradation in performance given emphasis towards either end and observe $150$ batches as a reliable amount. This contradicts the observation made in ITOP\cite{liu2021we} which favors smaller RigL update intervals for more updates. Additionally, setting $J$ and $Q$ to $1$, \textit{i.e.} without our Accuracy Improvement and Reactivate \& Explore stages, reduces results to RigL's performance ($73.1\%$ vs. $73.0\%$ for RigL~\cite{evci2020rigging}).

\noindent\textbf{Freezing $\Theta_K$ in \emph{Reactivate \& Explore} and inclusion of the \emph{Accuracy Improvement} stage.} We mentioned in Sec.\ref{subsec:lookahead_method} that the inclusion of \emph{Accuracy Improvement} stage and freezing currently selected architecture $\Theta_K$ in \emph{Reactivate \& Explore} are crucial to the performance with more thorough exploitation and stable exploration, which is now demonstrated and validated by the ablation results in Table~\ref{table:ablation}. We observe a significant drop in performance when we do not enable these features.

\noindent\textbf{Growing criterion and initialization.} We further study the sensitivity of our algorithm to the growing criterion and the initialization of the grown neurons. As shown in Table~\ref{table:ablation}, randomly growing the connections as in SET~\cite{mocanu2018scalable}(v.s. growing by the given importance score such as magnitude criterion after \emph{Reactivate \& Explore}) leads to NaN with overflowed gradients after a few epochs. If we use ZeroInit(v.s. MRU), that is, initializing the grown neurons to $0$ as in RigL~\cite{evci2020rigging} instead of inhering their most recently used values, the performance also drops. We achieve the best results with our proposed settings. 

\begin{table}[!t]
\centering
\begin{adjustbox}{width=\columnwidth}
\begin{tabular}{c|c|c|c|c|c}
    \toprule
    \small{Grow Criterion} &
    \small{Init} & \small{Freeze} & \small{\emph{Acc. Improv.}} & \small{Update Period (H, J, Q)} & \small{Top1 Acc($\%$)}$\uparrow$ \\
    \midrule
    \small{Magnitude} & \small{MRU} & \cmark & \cmark & $100, 100, 100$ & $73.9$ \\
    \rowcolor{lgreen} \small{Magnitude} & \small{MRU} & \cmark & \cmark & $150, 150, 150$ & $\mathbf{74.3}$ \\
    \small{Magnitude} & \small{MRU} & \cmark & \cmark & $200, 200, 200$ & $74.0$ \\
    \hdashline
    \small{Magnitude} & \small{MRU} & \cmark & \cmark & $100, 100, 150$ & $73.7$ \\
    \small{Magnitude} & \small{MRU} & \cmark & \cmark & $150, 150, 100$ & $73.8$ \\
    \small{Magnitude} & \small{MRU} & \cmark & \cmark & $200, 200, 150$ & $74.0$ \\
    \small{Magnitude} & \small{MRU} & \cmark & \cmark & $100, \mathit{1}, \mathit{1}$ & $73.1$ \\
    \midrule 
    \small{Magnitude} & \small{MRU} & \xmark & \cmark & $150, 150, 150$ & $73.6$ \\
    \midrule
    \small{Magnitude} & \small{MRU} & \cmark & \xmark & $150, 0, 150$ & $73.4$ \\
    \midrule
    \small{Random} & \small{MRU} & \cmark & \cmark & $150, 150, 150$ & $\text{NaN}$ \\
    \small{Magnitude} & \small{ZeroInit} & \cmark & \cmark & $150, 150, 150$ & $74.0$ \\
    \bottomrule
\end{tabular}
\end{adjustbox}
\caption{Performance of \method{} as a function of the update period, grown initialization, weight freezing, and inclusion of \emph{Accuracy Improvement} stage. Results with $90\%$ unstructured ERK sparsity on ImageNet1K and ResNet50 trained for 100 epochs.}
\label{table:ablation}
\vspace{-20pt}
\end{table}

\begin{figure}[!t]
\begin{center}
   \includegraphics[width=\linewidth]{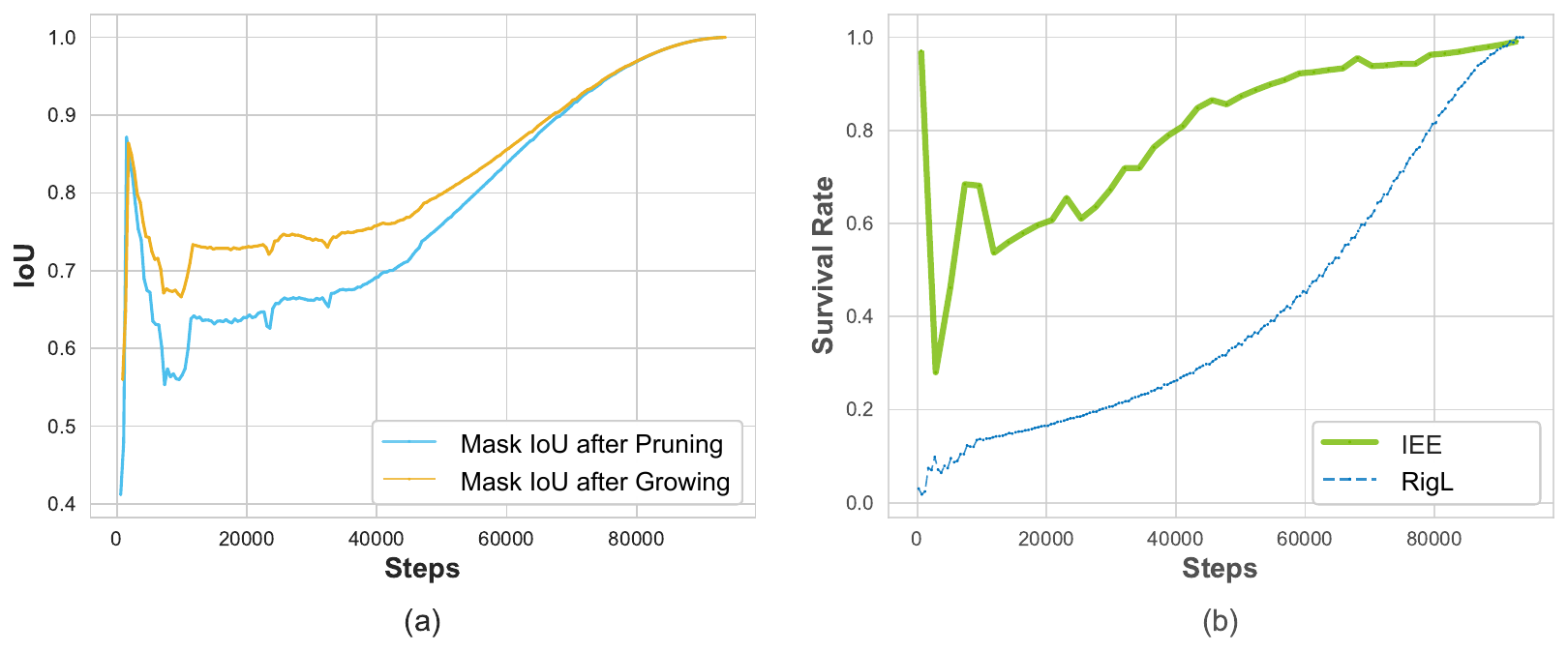}
\end{center}
    \vspace{-0.75cm}
   \caption{(a) Architecture convergence with IoU after pruning and growing; (b) Grown Neurons Survival Rate for ours and RigL~\cite{evci2020rigging}.}
   \label{fig:discussion}
   \vspace{-0.5cm}
\end{figure}

\subsection{Discussions}
\noindent\textbf{Architecture convergence.}
\label{subsec:discussion} We also analyze the convergence of the discovered sparse architecture. To this end, we design a convergence evaluation metric to compute the IoU of active sparse structure $\Theta_K$ for two consecutive \method{} step either after the \emph{Prune} or after the \emph{Grow} phase.

Figure~\ref{fig:discussion}(a) shows the evolution of the IoU as the training progresses as a proxy for architecture convergence. As we can see, for both pruning and growing sparse structure IoU there is a clear convergence towards $\text{IoU}=1$, indicating that the discovered sparse architecture becomes more stable towards the end of the training process.

\noindent\textbf{Effectiveness of the exploration strategy.}  
We hypothesized that the discrepancy in the criteria for pruning and growing in dynamic sparse training methods leads to a suboptimal exploration of new architectures, where newly grown parameters are mostly pruned before being fully exploited. To test this hypothesis and demonstrate the effectiveness of \method{} exploration strategy compared to the greedy strategies adopted in dynamic sparse training techniques like RigL\cite{evci2020rigging}, we propose a new metric named neuron growth \textbf{\textit{survival rate}}. The main idea is to gauge the fraction of newly grown neurons that are still active after the subsequent pruning step, indicating their reliability and usability once grown, joined with a side benefit to hint at architectural stability. Intuitively, a high survival rate suggests an effective growth step as those newly grown parameters persist in the architecture for future training. As shown in Figure~\ref{fig:discussion}(b), the survival rate of our approach is significantly higher than RigL~\cite{evci2020rigging} for all exploration steps, suggesting a more solid sparsity exploration.

\section{Conclusions}
\label{sec:concls}
In this paper, we present a novel approach to improve importance criteria for both unstructured and structured sparsity. We divide the model into an active structure for exploitation and an exploration space for potential updates. During exploitation, we optimize the active structure, whereas in exploration, we reevaluate and reintegrate parameters from the exploration space. Tested across various datasets and configurations, our approach demonstrates superior performance and cost efficiency compared to existing methods. We also perform a thorough ablation study and use specific metrics to assess the effectiveness of our exploration strategy.

\bibliographystyle{ieee_fullname}
\bibliography{main}

%
%
\clearpage
{\noindent \Large \textbf{Appendix}}
\setcounter{section}{0}

\section{Difference from DST Methods (e.g. RigL)}
Dynamic sparse training(DST) methods~\cite{mocanu2018scalable, dai2019nest, evci2020rigging, liusw2021sparse, liu2021we,yuan2021mest, yuan2022layer} like RigL perform growing with \emph{\textbf{single}} mini-batch of data $B$ by ranking \emph{sparse} gradients over $\Theta_P$: 
$$\frac{\partial}{\partial \Theta_P} \sum_{i=1}^{|B|}\ell(f(\Theta_K \cup \Theta_P; x^i), y^i) |_{\Theta_P = \mathbf{0}}$$
this greedy technique only cares effectiveness of growing for immediate next gradient descent step. In contrast, since our importance criterion is consistent in both \emph{Prune} and \emph{Grow}, combining our \emph{Reactivate \& Explore} and \emph{Grow} stages, our growing criterion could be considered as leveraging both \emph{\textbf{``prior"}} importance information and performing \emph{\textbf{``posterior"}} correction and adjustment based on newly selected $\Theta_K$ in the current \method{} update step. For example with magnitude criterion and $Q=1$, criterion of \method{} can be reformulated as:
$$Prior(|\Theta_P|)+\frac{\partial}{\partial \Theta_P} \sum_{i=1}^{|B|}\ell(f(\Theta_K\cup\Theta_P; x^i), y^i)$$, 
which considers importance from prior weights of $\Theta_P$ and also its adjusted weights in the \emph{Reactivate \& Explore} stage based on a new $\Theta_K$ selected in the current \method{} step.

This offers another perspective why our method could effectively reduce greediness in exploring new sparse architectures than others. Quantitative comparisons between our growing and the previous are also provided in \textbf{Sec.4.4} of the main paper.

\section{Ampere Pruning Results}
With the introduction of the NVIDIA Ampere GPU, researchers in the community began to consider leveraging ampere sparsity for acceleration and compression. With $N:M$ sparsity, we sparsify $N$ neurons out of $M$ contiguous neurons. With a $2:4$ ampere sparsity, the total number of parameters in the network will be halved, but this will be slightly more structured than a non-uniformly sparsified $50\%$ sparsity network and thus enjoy acceleration and computation saving. The proposed scheme of \method{} can also be instantiated in ampere pruning scenario. In Table \ref{tab:ampere}, we compare \method{} with three strong latest \emph{specialized(i.e. designed only for)} ampere pruning methods~\cite{mishra2021accelerating, zhou2021learning, lasby2023dynamic} and found that \method{} beats them in Top-1 by a margin with the same $2:4$ sparsity and training FLOPs needed. For example, compared with SR-STE\cite{zhou2021learning}, \method{} improves the Top-1 by $0.5$ with same $2:4$ sparsity and $\times0.83$ training cost. Compared with the latest SRigL~\cite{lasby2023dynamic}, we surpass its performance by almost $1$ point in Top-1 ($\mathbf{77.5}$ v.s. $76.6$). These results again validate the generalizability of our proposed \method{} scheme in ampere sparsity.    

\begin{table}[t!]
    \centering
    \resizebox{\linewidth}{!}{
    \begin{tabular}{l|cccc}
        \hline
        \rowcolor{lgray}
         \textsc{Method} & \textsc{Top-1 Acc}($\%$)$\uparrow$ &\textsc{N:M} & \textsc{Train FLOPs}($\times e^{18}$)$\downarrow$
         \\\hline
         \textsc{ASP~\cite{mishra2021accelerating}} & $76.8$ &$2:4$ & $\times 1.61$\\
         \textsc{STE~\cite{zhou2021learning}} & $76.4$ &$2:4$ & $\times 0.83$\\
         \textsc{SR-STE~\cite{zhou2021learning}} & $77.0$ &$2:4$ & $\times 0.83$\\
         \textsc{SRigL~\cite{lasby2023dynamic}} & $76.6$ &$2:4$ & $\times 0.83$\\
         \rowcolor{lgreen}
         \textsc{Ours} & $\mathbf{77.5}$ &$2:4$ & $\times 0.83$\\
         \hline
    \end{tabular}
    }
    \caption{\textbf{ImageNet1K} N:M sparsity results with ResNet50. \method{} surpasses strong latest specialized ampere pruning methods in Top-1 with the same training FLOPs needed.}
    \label{tab:ampere}
    \vspace{-0.5cm}
\end{table}

\section{Training FLOPs Computation}
In tables presented in the paper, we demonstrate the training cost of \method{} as well as other methods. FLOPs needed for a single forward pass inference of sparse model is computed by counting the total number of multiplications and additions. However, during training, the FLOPs computation would be slightly different due to different usage of the back-propagation gradients. In summary, training a neural network consists of $2$ main steps which are \emph{forward pass} and the \emph{backward pass}. During the \emph{forward pass}, we calculate the loss of the given batch of data using the current set of model parameters. Activations of each layer are stored in memory for the following backward pass. During the \emph{backward pass}, we use the loss value as the initial error signal and back-propagate the error signal to calculate the gradients of parameters. We calculate respectively the gradient of the activations of the previous layer and the gradient of its parameters. Roughly, the FLOPs needed for backward pass will be \textbf{\emph{twice}} the FLOPs needed for forward pass. Suppose a given dense architecture has forward pass FLOPs represented as $\zeta_D$ and its pruned or sparsified model has FLOPs $\zeta_P$. Training a sample with dense model can be expressed as $3\cdot\zeta_D$. 
\\\textbf{\textsc{\method{}}} Each \method{} step consists of three training stages, namely: \emph{Importance Estimation}, \emph{Accuracy Improvement}, and \emph{Reactivate \& Explore}. For each \emph{Importance Estimation} and \emph{Accuracy Improvement}, we need $3\times\zeta_P$ FLOPs for both sparse forward and backward pass. For \emph{Reactivate \& Explore}, since we are training with temporarily reactivated $\Theta_P$, we need $2\times\zeta_P + \zeta_D$ FLOPs to take care of the dense forward pass. We still use sparse gradients for updating due to the frozen $\Theta_K$. After the entire \method{} update period, the FLOPs needed would simply be $3\times\zeta_P$. Since the update period ends at $3/4$ of the entire training epochs, the average training cost can be calculated as:
$$\frac{3}{4}\cdot\frac{(H+J)\cdot3\cdot\zeta_P + Q\cdot(2\cdot\zeta_P + \zeta_D)}{H+J+Q} + \frac{1}{4}\cdot3\cdot\zeta_P$$
With $H=J=Q$, the cost would be: $$\frac{11\cdot \zeta_P+\zeta_D}{4}$$ This would be slightly higher than completely training a sparse model from scratch which is $3\cdot\zeta_P$ but still substantially lower than dense model training cost ($3\cdot \zeta_D$). \\
Also notice that, according to our above description of \method{} with structured sparsity, we follow the exponential scheduler of HALP~\cite{shen2021halp}, and the update period ends much earlier than $3/4$ of the total training epochs. The update with \method{} for latency-constrained structured sparsity will instead end at the $5$th epoch. The average training cost of \method{} will also be much lower. With $130$ training epochs in total, according to the calculation we provide above, it will instead be:
$$\frac{5}{130}\cdot\frac{(H+J)\cdot3\cdot\zeta_P + Q\cdot(2\cdot\zeta_P + \zeta_D)}{H+J+Q} + \frac{125}{130}\cdot3\cdot\zeta_P$$
With $H=J=Q$, the cost would approximately be:
$$\frac{388.3\cdot\zeta_P+1.7\cdot\zeta_D}{130}$$
\\\textbf{\textsc{Soft Masking}} Now for the family of soft masking methods like SNFS~\cite{dettmers2019sparse}, DPF~\cite{lin2020dynamic}, and DCIL~\cite{kim2021dynamic}, training cost vary based on different methods. Since these methods typically maintain dense gradients during backpropagation, training cost would usually be noticeably higher than typical sparse training approaches. For SNFS~\cite{dettmers2019sparse}, the total number of training FLOPs scales with $2\cdot \zeta_P+\zeta_D$. For DCIL~\cite{kim2021dynamic}, the work requires two forward and backward passes each time to measure two sets of gradients(one with dense weight and one with sparse weight) for weights update, and the total number of training FLOPs scales with $5\cdot \zeta_D + \zeta_P$, which is nearly doubled dense model training cost ($6\cdot\zeta_D$).
\\\textbf{\textsc{Zero-shot Pruning}} For the family of static sparse training or zero-shot pruning, the cost can be expressed as $3\cdot\zeta_P$ for both sparse forward and backward pass.
\\\textbf{\textsc{Pruning from Pretrained}} Most of the pruning from pretrained methods nowadays employed iterative pruning. For simplicity here, we estimate a \emph{very loose theoretical lowerbound} assuming one-shot pruning and no further gradients calculation on the pruned parameters during finetuning. The training cost of pretrained dense model scales with $3\cdot\zeta_D$ as discussed. In the later finetuning stage, the cost would scale with $3\cdot\zeta_P$ since the model deals with a sparse model now.
\\\textbf{\textsc{RigL~\cite{evci2020rigging}}} For the representative state-of-the-art dynamic sparse training work RigL, iterations with no connections updates need $3 \cdot \zeta_P$ FLOPs. At every $\Delta T$ iteration, RigL calculates the dense gradients. The averaged FLOPs for RigL is given by $\frac{3\cdot\zeta_P+2\cdot\zeta_P+\zeta_D}{\Delta T+1}$.
\\\textbf{\textsc{GraNet~\cite{liusw2021sparse}}} The difference between GraNet and RigL is at the starting sparsity of the method. RigL, same as our \method{}, starts from a sparse model of the target sparsity; whereas for GraNet, they start from a denser model of smaller sparsity. The best result from their paper, also shown in our main paper, starts at $50\%$ ERK sparsity($5.8$ FLOPs). However, the reported training FLOPs does not take the denser model pretraining into account. We explain here how we correct the training FLOPs calculation. In the first $30$ epochs, as described by GraNet~\cite{liusw2021sparse}, they gradually prune to the target sparsity, and the final model has $3.0$ FLOPs. We compute the average FLOPs in the first $30$ epochs simply as $(3.0 + 5.8) / 2 = 4.4$ and use it as $\zeta_P$ for the first $30$ epochs and $3.0$ as $\zeta_P$ for the remaining training epochs.
\\\textbf{\textsc{MEST, SpFDE~\cite{yuan2021mest, yuan2022layer}}} We just use the reported training FLOPs in the paper. However, notice that these two methods, besides sparse training, also leverage orthogonal augmentation techniques like data sieving and layer freezing to additionally reduce training costs. In \method{}, we only perform sparse training as in RigL and others for a fair comparison.
\\\textbf{\textsc{Interspace Pruning}} For the very latest interspace pruning work~\cite{wimmer2022interspace}, authors use FB convolution layers which introduce additional forward and backward overhead. Given the information provided in the paper, for a particular convolution layer with size $c_{out}\times c_{in}\times K \times K$, the relative increase of forward pass would be $K^2/c_{out}$ times the dense forward pass. Notice that this is a constant overhead independent of the pruning rate and sparsity of the model. Similarly, the authors provide that the backward pass would introduce an additional constant overhead of $K^2/c_{in}$ times the dense computation of gradients. Since authors provide no exact FLOPs of the model, we also estimate a lower bound of $K^2/c_{out}$ and $K^2/c_{in}$ as $3^2/128 \approx 0.07$ for ResNet-50. This is a lower bound since as identified in many works before the spatial size is the largest in the early layers with a large $K$ and small $c_{out}$ and $c_{in}$ processing large-sized feature maps and dominating the overall FLOPs of the model. Now we could calculate the FLOPs needed to train a single example as $\zeta_P + 0.07\cdot\zeta_D + 2\cdot(\zeta_P + 0.07\cdot\zeta_D)$ which is approximately $3\cdot\zeta_P + 0.21\cdot\zeta_D$.
\\\textbf{\textsc{NAS-based Methods}} We also demonstrate the results of some NAS-based methods~\cite{liu2019metapruning, dong2019network, guo2020dmcp} in the main paper table for comparison. Since the searching involved is very hard to quantify the training cost estimation, we only report the estimated training cost of the discovered pruned model calculated as ($3\cdot\zeta_P$). Now notice that this is a very loose lower bound, and the actual cost could be much higher with the architecture search.

\section{Integration of \method{} with Latency-Constrained Structured Sparsity}
We now present \method{} with latency-constrained structured sparsity setting. Specifically, we will highlight how we integrate with the latest latency pruning method HALP~\cite{shen2023hardware}.\\

\subsection{Recap of HALP and Latency-Constrained Pruning}
\label{subsubsec:halp_recap}
For our latency-constrained structured sparsification, we follow the latest resource-constrained pruning method HALP~\cite{shen2021halp} but impose the dynamic regime of \method{} to enhance the quality of the pruned model structure. Same as HALP, we formulate the pruning step as a global cost-constraint importance maximization problem, where we take into account the latency benefits incurred every time we remove or grow a channel from one of the layers of the network. Similarly, we also formulate our unique growing part as a cost-constraint importance maximization problem. In this section, we will provide a brief recap of HALP and how it's used our \method{} iterative prune-and-grow setup. Given a global resource constraint $\Psi$ defining the maximum amount of resource we could use, HALP aims to find a set of channels defining a sub-network achieving the best performance under the constraint $\Psi$. In this case, $\Psi$ represents the inference latency for a target hardware platform, for example the Nvidia TitanV GPU. 

HALP then prepare a latency lookup table $\mathcal{T}$, where $\mathcal{T}^l(p^{l-1}, p^l)$ records the layer latency at layer $l$ with $p^{l-1}$ active input channels and $p^l$ active output channels. With this latency look-up table, HALP associates a potential latency reduction value $R_j^l$ to each $j\text{th}$ channel of layer $l$, computed as follows:
\begin{align}
    R_j^l = T^l(p^{l-1}, j) - T^l(p^{l-1}, j-1), 1 \leq j \leq p^l
\end{align}
$R_j^l$ estimates the potential latency saving if we prune the corresponding channel. Now, in order to estimate the performance of the selected sub-newtwork, HALP measures the importance score $\mathcal{I}^l_j$ for each $j\text{th}$ channel of layer $l$. The importance score metric adopted here is Taylor importance~\cite{molchanov2019importance}, which is evaluated as follows:
\begin{align}
    \mathcal{I}^l_j = |g_{\gamma^l_j}\gamma^l_j + g_{\beta^l_j}\beta^l_j|,
\end{align}
where $\gamma$ and $\beta$ are the BatchNorm layer's weight and bias. With $R$ and $\mathcal{I}$ calculated, HALP formulates the latency-constrained channel pruning as a Knapsack problem where we try to maximize the total importance but under the latency constraint $\Psi$. The pruned channels can then be selected by an augmented Knapsack solver $Knapsack(\mathcal{I}, R, \Psi)$, which returns the items achieving maximum importance while the accumulated latency cost is below the global constraint $\Psi$.  

\subsection{IEE Prune Step}
As described in the main paper, in the \emph{Prune} phase of the $t$-th \method{} step, we are going to prune a number of parameters that satisfy the ``update budget" $\Omega^t$. For integration with the HALP framework, we first collect the importance $\mathcal{I}$ and channel latency cost $R$ from the \textbf{active sparse structure} $\Theta_K$. Since the initialized compute resource is $\Psi$, after the Prune step, our desired target is $\Psi - \Omega^t$. We then leverage the Knapsack solver $Knapsack(\mathcal{I}, R, \Psi- \Omega^t)$ to choose which channels to transfer from $\Theta_K$ to $\Theta_P$.

\subsection{IEE Grow Step}
During growing, we also want to take the model latency into account to prevent some latency-costly channels from getting added back. We perform a similar latency-constrained selection as above. Here, we collect the importance $\mathcal{I}$ and channel latency cost $R$ from the \textbf{exploration space} $\Theta_P$ after our \emph{Reactive \& Explore} step. Then, we grow a number of parameters that satisfy the ``update budget" $\Omega^t$. Channels selected by the Knapsack solver $Knapsack(\mathcal{I}, R, \Omega^t)$ are transferred from $\Theta_P$ to $\Theta_K$.

\section{Detailed Experiment Hyperparameter and Optimization Settings}
The large-scale image classification dataset ImageNet~\cite{deng2009imagenet} is of version ILSVRC2012~\cite{russakovsky2015imagenet}, which consists of $1.3M$ images of $1000$ classes. We run all experiments on ImageNet and PASCAL VOC with eight NVIDIA Tesla V100 GPUs. Experiments on CIFAR10~\cite{krizhevsky2009learning} are conducted with a single NVIDIA Tesla V100 GPU. All experiments are conducted with PyTorch~\cite{paszke2017automatic} V1.4.0. 
\subsection{Layerwise Sparsity Distribution}
In the main paper, in our unstructured sparsity setting, we demonstrated results of \method{} and comparison with three types of sparsity distribution, namely: Uniform, ERK, and Non-Uniform. Here, to clear possible confusion, we provide more detailed explanations here. Given a predefined sparsity $S$ or a number of available neurons, we have different ways to allocate them across layers, which also results in different FLOPs. With Uniform sparsity, the sparsity of each layer $S^{l}$ is equal to the total sparsity $S$ throughout training, i.e., $S = S^{l}$. With ERK sparsity, we use the \textit{Erd\H{o}s-R\'enyi-Kernel} (ERK) formulation~\cite{mocanu2018scalable, evci2020rigging} to set sparsity for each layer, which means higher sparsity is assigned to those layers with more parameters i.e., $S^{l_i} > S^{l_j}$ if $m^{l_i} > m^{l_j}$, where $m^{l_i}$ represents the number of parameters for layer $l_i$. With Non-Uniform sparsity, we do not pose any constraints for layerwise distribution. Concretely in pruning, with Non-Uniform sparsity, we simply rank all neurons in the model globally. Though all Uniform, ERK, and Non-Uniform are \emph{unstructured} sparsity, Uniform is slightly more \emph{structured} than ERK which is also naturally more \emph{structured} than Non-Uniform.
\subsection{Unstructured Weight Sparsity on ResNet50-ImageNet}
\label{subsubsec:hypweightres50}
We use an individual batch size of $128$ per GPU and follow NVIDIA's recipe ~\cite{nvidia2020} with mixed precision and Distributed Data Parallel training. The learning rate is warmed up linearly in the first $8$ epochs reaching its highest learning rate then follows a cosine decay~\cite{loshchilov2016sgdr} over the remaining epochs. The pretrained model weight is kept consistent with RigL~\cite{evci2020rigging} to ensure a fair comparison.

\subsection{Unstructured Weight Sparsity on WideResNet22-2-CIFAR10}
In our experiments section, we also include results of WideResNet22-2, which is Wide Residual Network ~\cite{zagoruyko2016wide} with $22$ layers using a width multiplier of $2$. We use an individual batch size of $128$, an initial learning rate of $0.1$ decaying by a factor of $5$ every $30000$ iterations, an L2 regularization coefficient of $5e-4$, and a $SGD$ momentum of $0.9$. Similarly, results of RigL are reproduced using the same hyperparameters and optimization settings as ours, ensuring a fair comparison. 
\subsection{Latency Constrained Structured Sparsity}
\textbf{ImageNet} We follow HALP~\cite{shen2021halp} for setting the hyperparameters and optimization settings of experiments on latency constrained structured sparsity with ResNet50 and MobileNet-V1. They are also similar to the recipe described in \ref{subsubsec:hypweightres50}. We also follow HALP~\cite{shen2021halp} for constructing the latency lookup table, which is pre-generated targetting the NVIDIA TITAN V GPU inference by iteratively reducing the number of channels in a layer and characterize the corresponding latency with NVIDIA cuDNN~\cite{chetlur2014cudnn} V7.6.5. The latency measurement is conducted $100$ times to avoid randomness. We also refer to HALP for some special implementation detail such as how to deal with the group convolution in MobileNet-V1, negative latency contribution, pruning of the first model layer, which are all described in detail in HALP.\\
\textbf{PASCAL VOC} We follow the "07 + 12" setting as in \cite{liu2016ssd} and use the union of VOC2007 and VOC2012 trainval as our training set and VOC2007 test as test set. Our SSD model, similar to HALP~\cite{shen2021halp}, is based on \cite{liu2016ssd}. Following \cite{huang2017speed}, for efficiency, we remove the last stage of convolution layers, last avgpool, and fc layers from the original ResNet50 classification structure. Also, all strides in the third stage of ResNet50 layer are set to $1\times1$. We train our models for $900$ epochs with SGD optimizer and learning rate schedule same as \cite{shen2021halp} with an initial learning rate of $8e-3$ which warms up in the first $50$ epochs then decays by $3/8, 1/3, 2/5, 1/10$ at the $700, 800, 840, 870$th epoch

\end{document}